\documentclass[10pt,twocolumn,letterpaper]{article}

\usepackage{cvpr}              

\usepackage[dvipsnames]{xcolor}

\definecolor{cvprblue}{rgb}{0.21,0.49,0.74}
\usepackage[pagebackref,breaklinks,colorlinks,citecolor=cvprblue]{hyperref}
\usepackage{amsmath}
\usepackage{mathrsfs}
\usepackage{mathalpha}
\usepackage{caption}
\usepackage{subcaption}

\def\paperID{*****} 
\def\confName{CVPR}
\def\confYear{2024}

\title{Short-term Object Interaction Anticipation with Disentangled Object Detection \\ @ Ego4D Short Term Object Interaction Anticipation Challenge}

\author{
Hyunjin Cho$^{1}$,\hspace{1cm} Dong Un Kang$^{1}$,\hspace{1cm} Se Young Chun$^{1,2,}$\thanks{Corresponding author}\\
$^{1}$Dept. of ECE, $^{2}$INMC \& IPAI, Seoul National University, Republic of Korea\\
\texttt{\{jim0228, qkrtnskfk23, sychun\}@snu.ac.kr}
}

\begin{document}
\maketitle
\begin{abstract}
Short-term object interaction anticipation is an important task in egocentric video analysis, including precise predictions of future interactions and their timings as well as the categories and positions of the involved active objects. To alleviate the complexity of this task, our proposed method, SOIA-DOD, effectively decompose it into 1) detecting active object and 2) classifying interaction and predicting their timing. Our method first detects all potential active objects in the last frame of egocentric video by fine-tuning a pre-trained YOLOv9. Then, we combine these potential active objects as query with transformer encoder, thereby identifying the most promising next active object and predicting its future interaction and time-to-contact. Experimental results demonstrate that our method outperforms state-of-the-art models on the challenge test set, achieving the best performance in predicting next active objects and their interactions. Finally, our proposed ranked the third overall top-5 mAP when including time-to-contact predictions. The source code is available at \url{https://github.com/KeenyJin/SOIA-DOD}.
\end{abstract}


\section{Introduction}
\label{sec:intro}
Short-term object interaction anticipation is a complex computer vision task that aims to predict human interactions with objects in the near future and estimate the timing of these interactions from an egocentric perspective (\textit{i.e.,} the viewpoint of the person). The Ego4D dataset~\cite{grauman2022ego4d} is a collection of massive and diverse egocentric videos to provide several benchmark challenges for understanding the first-person visual experience in the past, present and future. Here, we focus on the short-term anticipation task, specifically forecasting future human-object interaction. Successfully predicting the next human actions would enable robots and augmented reality systems to better assist and protect humans.
However, the requirement for accurate forecasting in both spatial (location) and temporal (timing) domains makes short-term object interaction anticipation particularly challenging.

Recently, several works have explored anticipating human behavior in the most recent frame of the video clip using Ego4D dataset~\cite{grauman2022ego4d}.
StillFast~\cite{ragusa2023stillfast} introduced an end-to-end model that utilizes a two-branch architecture to process the high-resolution last observed frame and the low-resolution past frames simultaneously, fuses the spatial and temporal features, and then predicts the next active objects, actions, and time-to-contact.
GANOv2~\cite{thakur2023guided}, following the two-branch architecture of StillFast, employs a guided-attention mechanism to fuse spatial and temporal features and object detections from the video.
TransFusion~\cite{pasca2023summarize} proposed multimodal fusion, integrating language summaries of the video with the visual features of the last observed frame. While existing methods focused on simultaneously handling with spatial and temporal features for predicting the objects, actions, and time together, effectively training a model for these different multi-task learning aspects remains challenging, particularly for next active object detection, interaction classification, and timing prediction in the near future.

In this work, we propose a Short-term Object Interaction Anticipation with Disentangled Object Detection (SOIA-DOD) to address the challenge of multi-task learning by decomposing the task into two cascaded stages: active object detection, and prediction of interaction and its occurrence time. This approach alleviates the difficulty of learning both tasks simultaneously, and enables accurate localizing the potential active objects. The detected results of these objects are then used as initial queries and combined with a transformer encoder to select the most likely next active object, classify the interaction, and predict its timing.
In the Ego4D Short-Term Object Interaction Anticipation challenge, we demonstrate the effectiveness of our proposed SOIA-DOD, achieving state-of-the-art performance in predicting the next active objects and their interactions. Furthermore, it achieved the third place in the overall top-5 mean average precision (mAP) including time-to-contact.

\begin{figure*}[!t]
    \centering
    \includegraphics[width=1.00\textwidth]{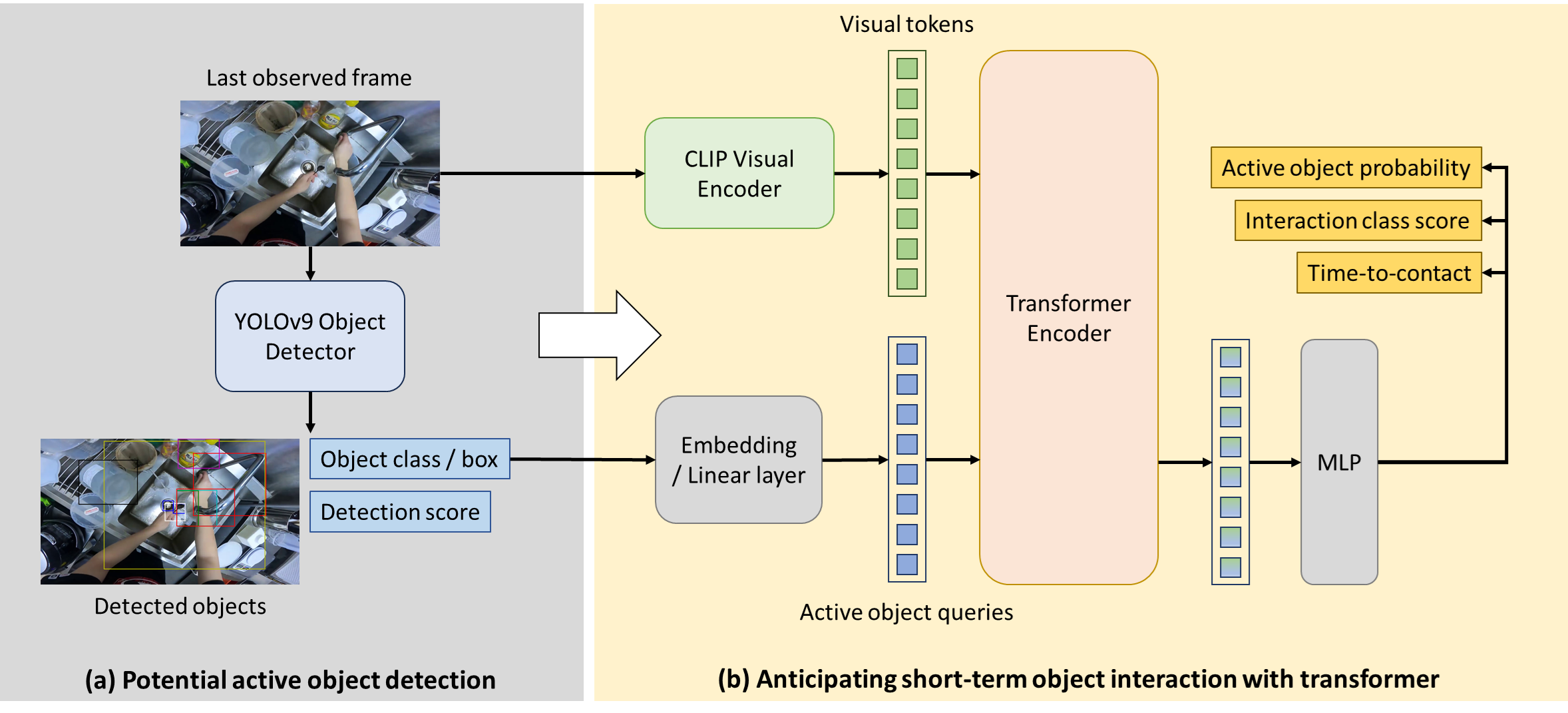}
    \caption{\textbf{The overall pipeline of our proposed SOIA-DOD.} The SOIA-DOD consists of two cascaded stages. (a) Potential active object detection: A disentangled object detection model (fine-tuned YOLOv9) detects potential active objects from the last egocentric frame. Based on the detection confidence scores, it selects the top-k active objects and generates queries representing these objects, including their class labels and bounding boxes.
    (b) Anticipating short-term object interaction with transformer: Visual tokens and active object queries are concatenated and processed by the self-attention mechanism within a transformer layer. This allows the model to learn relationships between different parts of the image and the active objects. Finally, the model outputs final predictions, including active object probability, interaction class score and time-to-contact.}
    \label{fig:pipeline}
\end{figure*}

\section{Method}
\label{sec:method}

In this section, we introduce our proposed SOIA-DOD, which anticipates future object interaction from an egocentric video. Our method consists of two cascaded stages: 1) Detection of potential active objects and 2) Selection of the most likely active object, along with predicting each action and its time-to-contact. The overall process of our SOIA-DOD is visualized in the Figure.~\ref{fig:pipeline}.

\subsection{Potential Active Objects Detection}
The first stage of our SOIA-DOD aims to detect potential active object present in the image. Using the Ego4D dataset, we fine-tune a pre-trained real-time object detector YOLOv9~\cite{wang2024yolov9} to estimate the location and class of next active object. Taking the last frame of given egocentric video, the fine-tuned YOLOv9~\cite{wang2024yolov9}, which is an active object detection model, outputs bounding box $b_i$, class label $c_i$, and detection score $\sigma_i$ for all potential active objects $\{O_i\}_{i=1}^{N}$. $N$ denotes the number of predictions.

Before proceeding to the next stage, we select top-k active object candidates based on the detection score $\sigma_i$. We then encode the bounding boxes $b_i$ and class names $c_i$ of these selected objects separately. The bounding boxes are encoded by single linear layer ($B \in R^{k\times d}$), while the class names are encoded by embedding layer ($C \in R^{k\times d}$) where $d$ denotes the embedding dimension. Lastly, we construct the active object query $Q \in R^{k\times d}$ by summing the embeddings of bounding box $B$ and class names $C$.

\subsection{Anticipating Short-term Object Interaction with Transformer}
In the second stage, we identify the most likely next active object along with its interaction and time-to-contact predictions by combining the visual features and active object query $Q$ from the last egocentric frame. For the visual feature extraction, we use the pre-trained CLIP visual encoder (ViT-L/14@336px)~\cite{radford2021learning}. As demonstrated in~\cite{liu2024visual}, we extract the grid feature before the last transformer layer taking the last egocentric frame for improved visual understanding. These visual features are projected through a linear layer, forming a visual token for each patch, as denoted as $V\in R^{T\times d}$ where $T$ refers to the number of CLIP visual tokens. Then, we concatenates the active object query $Q$ from the first stage and the visual features $V$ for constructing the integrated active object query $Q'\in R^{(k+T)\times d}$.

Next, we utilize a standard transformer encoder to fuse the visual features and potential active objects features. The initial query $Q'$ is projected to the layers of transformer encoder. In each layer, the query is updated through the self-attention mechanism, applying layer normalization before the attention and feed-forward operations, respectively. Then, the updated active object queries $\{Q''_i\}_{i=1}^{N}$ are processed in multiple linear layers to predict 1) the probability of next active object $p_{obj, i}$ via a linear layer with a sigmoid function, 2) the probability distribution of future interaction $p_{int, i}$ via a linear layer with a softmax function and 3) the time-to-contact via three linear layers with ReLU in between and a softplus activation at the end. In inference, we multiply each query's object detection score $\sigma_i$ and top-K interaction scores to compute the final prediction score $s_{i, j} = \sigma_i \times p_{int, i, j}$. Here, $i=1,\ldots,N$ denotes each detected object and $j=1,\ldots,K$ represents each interaction among the top-K interactions of the i-th object. Final prediction is determined by the highest score $s$.

\subsection{Training Objective}

We first fine-tune the pre-trained object detector and then train the rest of our model while freezing the weight of object detector.
For fine-tuning, we follow the training objective outlined in~\cite{wang2024yolov9} to detect potential active objects.
For training the rest of the model, we use three types of losses: a binary cross entropy \(\mathcal{L}_{\text{obj}}\) for future active object, a cross entropy \(\mathcal{L}_{\text{int}}\) for interaction prediction, and a smooth L1 loss \(\mathcal{L}_{\text{ttc}}\) for time-to-contact regression.

\begin{equation}
\mathcal{L}_{total} = \mathscr{\lambda}_{\text{1}}\mathcal{L}_{\text{obj}} + \mathscr{\lambda}_{\text{2}}\mathcal{L}_{\text{int}} + \mathscr{\lambda}_{\text{3}}\mathcal{L}_{\text{ttc}}
\label{eq:loss}
\end{equation}

We also incorporate auxiliary layer losses during training, inspired by ~\cite{carion2020end}. These losses provide additional supervision for the model. The loss $\mathcal{L}_{total}$ is calculated for each output from the second-to-last layer of the transformer.

\begin{table}[t]
    \centering
    \begin{tabular}{l|cccc}
        \hline
        Methods & Noun & N+V & N+TTC & Overall \\
        \hline
        StillFast~\cite{ragusa2023stillfast} & 25.06 & 13.29 & 9.14 & 5.12 \\

        GANOv2~\cite{thakur2023guided} & 25.67 & 13.60 & 9.02 & 5.16 \\

        Zarrio* & 33.50 & 17.26 & 11.77 & 6.75 \\

        EgoVideo* & 31.08 & 16.18 & \textbf{12.41} & \textbf{7.21} \\

        SOIA-DOD (ours) & \textbf{34.89} & \textbf{17.61} & 10.91 & 6.22 \\
        \hline
    \end{tabular}
    \caption{\textbf{Top-5 mAP (\%) results on the test set of Ego4D v2.} 
    We evaluated short-term object interaction anticipation methods on the Ego4D benchmark. Our SOIA-DOD model outperforms other models in predicting the noun and verb categories. N, V, and TTC denote noun, verb, and time-to-contact, respectively. * denotes the participant team name of the challenge. For each metric, we highlighted the best performance in bold.}
    \label{tab:leaderboard}
\end{table}

\section{Experiments}\label{sec:experiments}

\subsection{Implementation Details}
We use version 2.0 of the Ego4D dataset for training and validation. We employ YOLOv9-E as the object detector and utilize the image encoder of CLIP ViT-L/14@336px for visual feature extraction. The input image is resized to a width of 1024 pixels for object detection and to 336x336 pixels for visual feature extraction with CLIP.
First, the object detector is trained using the SGD optimizer for 16 epochs with a learning rate of 1e-3 and a batch size of 16.
Then, our model except for the weights of the object detector is trained using the AdamW optimizer for 12 epochs with a batch size of 32 and a weight decay of 1e-3.
The learning rate for CLIP is set to 1e-5 and the learning rate for the rest of the model is set to 1e-4, both decaying by a factor of 0.1 every 11 epochs.
We set \(\mathscr{\lambda}_{\text{1}}\)=2.0, \(\mathscr{\lambda}_{\text{2}}\)=2.0, and \(\mathscr{\lambda}_{\text{3}}\)=1.0 in the loss $\mathcal{L}_{total}$.
During training, we select top-10 active object candidates to construct active object query $Q$. At inference, we select top-20 active object candidates and top-6 interactions for each object.

\begin{table}[t]
    \centering
    \begin{tabular}{l|cccc}
        \hline
        The number of\\object candidates & Noun & N+V & N+TTC & Overall \\
        \hline
        5 & 30.14 & 14.54 & 9.219 & 4.91 \\

        10 & 30.65 & \textbf{15.22} & \textbf{9.222} & \textbf{4.98} \\

        20 & \textbf{30.94} & 14.88 & 8.86 & 4.87 \\
        \hline
    \end{tabular}
    \caption{\textbf{Ablation study on the number of potential active object candidates for transformer queries.} We report Top-5 mAP (\%) results on the validation set of Ego4D. N, V, and TTC denote noun, verb, and time-to-contact, respectively. For each metric, we highlighted the best performance in bold.}
    \label{tab:ablation}
\end{table}

\begin{figure*}[htbp]
    \centering
    \includegraphics[width=0.85\textwidth]{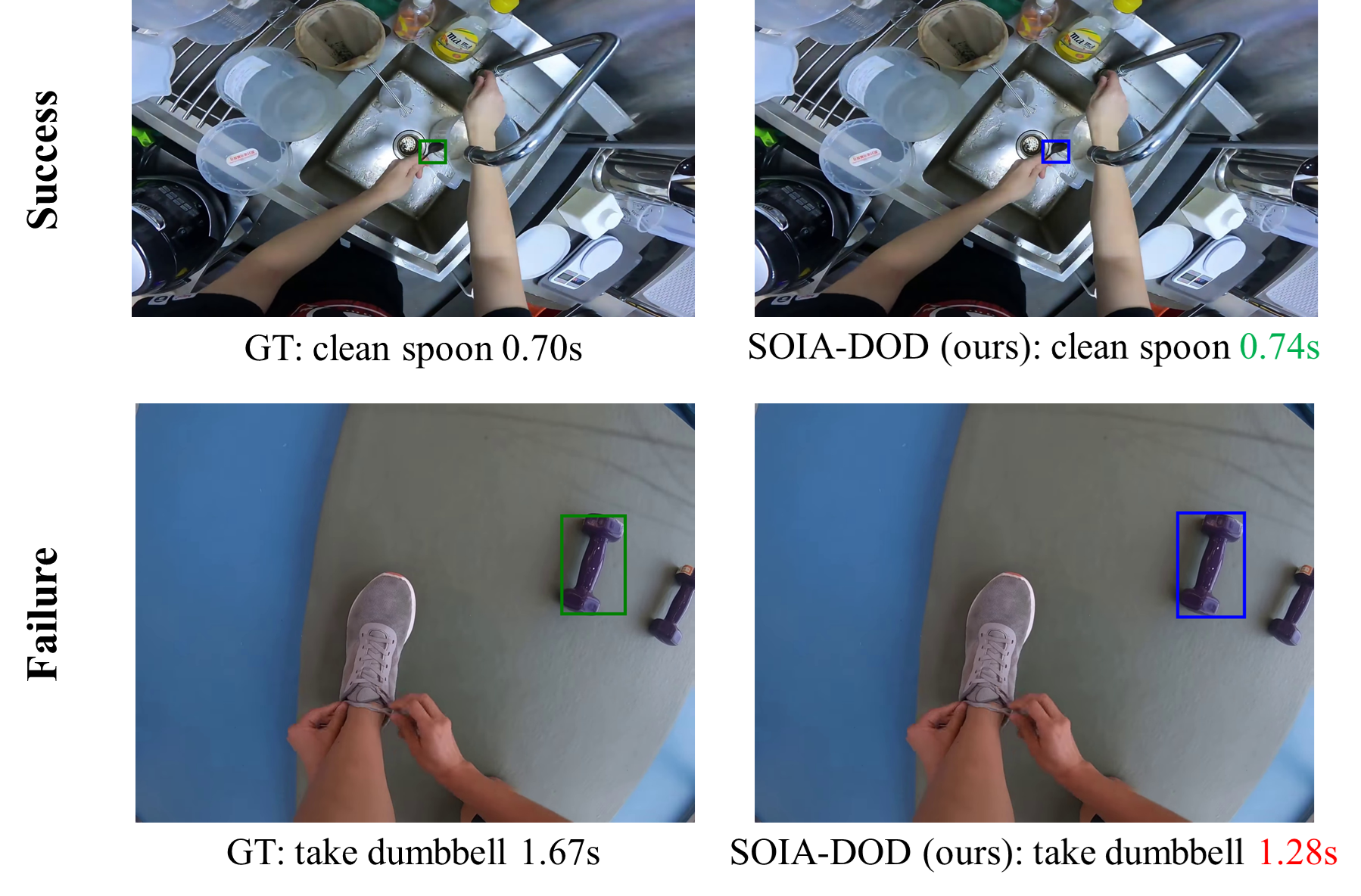}
    \caption{\textbf{Qualitative success and failure cases of our SOIA-DOD.} Our SOIA-DOD excels at detecting active objects and their interactions, but it often struggles with accurately estimating time-to-contact.}
    \label{fig:samples}
\end{figure*}

\subsection{Experimental Results}
We conduct the quantitative experiments on Ego4D Short-term Interaction Anticipation benchmark to demonstrate the effectiveness of our proposed SOIA-DOD. As shown in Table~\ref{tab:leaderboard}, our method achieves the third rank in the overall mAP for prediction of noun (next active object), verb (human interaction) and ttc (time-to-contact). While ranking third overall, it outperforms other methods in noun prediction by 1.39\% and in combined noun and verb prediction by 0.35\%.
In Table~\ref{tab:ablation}, we investigated the impact of the number of object candidates used to construct active object query $Q$ during training. 
Using the top-10 candidates yielded the best overall prediction performance during training.
We also visualize the success and failure cases of our proposed SOIA-DOD in Figure.~\ref{fig:samples}. 
While SOIA-DOD demonstrates accurate detection of active objects and their interactions, it often struggles with precisely estimating time-to-contact.

\subsection{Limitations}
While our proposed SOIA-DOD excels in predicting the next active objects and their interactions, it exhibits lower accuracy in forecasting time-to-contact compared to other state-of-the-art methods. This limitation likely stems from the model's reliance solely on last egocentric frame as input, consequently hindering its ability to leverage temporal information for precise time-to-contact prediction.

\section{Conclusion}\label{sec:conclusion}
We present a novel cascaded approach, SOIA-DOD, for short-term object interaction anticipation in egocentric videos. SOIA-DOD disentangles active object detection from the prediction of its human interaction and time-to-contact. By avoiding the challenges of simultaneously processing very different tasks, our method achieves accurate localized active object detection. In the short-term anticipation in Ego4D challenge, we demonstrates that SOIA-DOD outperforms state-of-the-art methods in predicting next active objects and their interactions, though it ranks third when including time-to-contact. Future work should focus on integrating more temporal information from past frames to enhance time-to-contact predictions.




{
    \small
    \bibliographystyle{ieeenat_fullname}
    \bibliography{main}
}
\end{document}